# On the Stability of Empirical Risk Minimization in the Presence of Multiple Risk Minimizers

Benjamin I. P. Rubinstein, Aleksandr Simma

*Abstract*—Recently Kutin and Niyogi investigated several notions of algorithmic stability—a property of a learning map conceptually similar to continuity—showing that training-stability is sufficient for consistency of Empirical Risk Minimization while distribution-free CV-stability is necessary and sufficient for having finite VC-dimension. This paper concerns a phase transition in the training stability of ERM, conjectured by the same authors. Kutin and Niyogi proved that ERM on finite hypothesis spaces containing a unique risk minimizer has training stability that scales exponentially with sample size, and conjectured that the existence of multiple risk minimizers prevents even super-quadratic convergence. We prove this result for the strictly weaker notion of CV-stability, positively resolving the conjecture.

## I. Introduction

DEVROYE and Wagner [3] first studied the effect of algorithmic stability on the statistical generalization of learning. Since then many authors have proposed numerous alternate notions of stability and have used them to study the performance of algorithms not easily analyzed with other techniques [1], [2], [4]–[6]. While the learnability of a concept class can now be characterized by the admission of a stable learner, finding natural definitions of stability with such properties is still open [5], [6]. As part of their general investigation into algorithmic stability in [5], Kutin and Niyogi observed that Empirical Risk Minimization (ERM) on hypothesis spaces of cardinality two experiences a phase transition in achievable rates of stability as the number of risk minimizers increases from one to two. Under a unique minimizer stability scales exponentially with sample size, while in the presence of multiple minimizers no rate faster than quadratic is possible. Kutin and Niyogi extended the unique risk minimizer result to finite spaces, and conjectured that ERM on a finite hypothesis space containing multiple risk minimizers is $(0,\delta)$-training-stable for no $\delta = o(m^{-1/2})$ [5, Conjecture 10.11]. We positively resolve this conjecture for weaker CV-stability and show furthermore that ERM on such spaces is $(0,\delta)$-CV-stable for $\delta = O(m^{-1/2})$.

Studying the effects of multiple risk minimizers is well motivated by the practical consideration of feature selection. Learning from irrelevant features can lead to multiple risk minimizers, for example when learning in a symmetric hypothesis class with positive minimum risk. Results showing poor generalization in such cases, as implied by slow rates of stability, help to justify the use of feature selection. It is also noteworthy that restricting focus to finite hypothesis classes is far from unreasonable—practitioners often attempt to learn on extracted features, which frequently involves discretization, and when learning on a finite domain any hypothesis space becomes effectively finite.

## II. Preliminaries

We follow the setting of [5] closely. $\mathcal{X}$ denotes the *input space*, $\mathcal{Y} = \{-1, 1\}$ the *output space* and $\mathcal{Z} = \mathcal{X} \times \mathcal{Y}$ their product *example space*. $D$ is a distribution on $\mathcal{Z}$. Unless stated otherwise we assume that $m$ *examples* are drawn i.i.d. according to $D$. A *classifier* or *hypothesis* is a function $h : \mathcal{X} \to \{-1, 1\}$; $\mathcal{H}$ denotes a set of classifiers or a *hypothesis space*. Whenever we refer to a finite $\mathcal{H}$ we assume, without loss of generality, that no $h_1, h_2 \in \mathcal{H}$ exist such that $h_1 \neq h_2$ and $h_1(X) = h_2(X)$ a.s.

A *learning algorithm* on hypothesis space $\mathcal{H}$ is a function $\mathcal{A} : \bigcup_{m>0} \mathcal{Z}^m \to \mathcal{H}$. When $\mathcal{A}$ is understood from the context, we write $f_s = \mathcal{A}(s)$ for $s \in \mathcal{Z}^m$. The *loss* incurred by classifier $h$ on example $z \in \mathcal{Z}$ is denoted by $\ell(h, z)$. We will assume the 0-1 loss throughout, where $\ell(h, (x, y)) = \mathbf{1}[h(x) \neq y]$. Define the *risk* of $h$ with respect to $D$ as $\mathrm{R}_D(h) = \mathbb{E}_{Z \sim D}[\ell(h, Z)]$; and with a slight abuse of notation the *empirical risk* as $\mathrm{R}_s(h) = m^{-1} \sum_{i=1}^m \ell(h, z_i)$ for $s = (z_1, \ldots, z_m) \in \mathcal{Z}^m$. The risk minimizers of space $\mathcal{H}$ with respect to distribution $D$ is the set $\mathcal{H}^\star = \arg\min_{h \in \mathcal{H}} \mathrm{R}_D(h)$. A learning algorithm $\mathcal{A}$ is said to implement Empirical Risk Minimization if $\mathcal{A}(s) \in \arg\min_{h \in \mathcal{H}} \mathrm{R}_s(h)$ for $s \in \mathcal{Z}^m$.

Based on sequence $s = (z_1, \ldots, z_m) \in \mathcal{Z}^m$, index $i \in [m]$ and example $u \in \mathcal{Z}$, we define sequences $s^i = (z_1, \ldots, z_{i-1}, z_{i+1}, \ldots, z_m) \in \mathcal{Z}^{m-1}$ and $s^{i,u} = (z_1, \ldots, z_{i-1}, u, z_{i+1}, \ldots, z_m) \in \mathcal{Z}^m$. The following definitions capture the relevant notions of stability.

*Definition 2.1:* A learning algorithm $\mathcal{A}$ is *weakly $(\beta, \delta)$-hypothesis-stable* or has *weak hypothesis stability* $(\beta, \delta)$ if for any $i \in [m]$,

$$\Pr_{(S,U) \sim D^{m+1}} \left( \max_{z \in \mathcal{Z}} |\ell(f_S, z) - \ell(f_{S^{i,U}}, z)| \leq \beta \right) \geq 1 - \delta \ .$$

The notion of weak hypothesis stability was first used by Devroye and Wagner in [3] under the name of *stability*. Kearns and Ron [4] then used the term *hypothesis stability* for the same concept.

Manuscript received February 10, 2010. This work was supported by the Siebel Scholars Foundation and the National Science Foundation under grant DMS-0707060.

The authors are with the Computer Science Division, University of California, Berkeley (e-mail: {benr,asimma}@eecs.berkeley.edu).

*Definition 2.2:* A learning algorithm $\mathcal{A}$ is $(\beta,\delta)$-*cross-validation-stable*, or $(\beta,\delta)$-*CV-stable*, if, for any $i \in [m]$,

$$\Pr_{(S,U) \sim D^{m+1}}(|\ell(f_S, U) - \ell(f_{S^{i,U}}, U)| \leq \beta) \geq 1 - \delta .$$

*Definition 2.3:* A learning algorithm $\mathcal{A}$ is $(\beta,\delta)$-*overlap-stable*, or has *overlap stability* $(\beta,\delta)$, if, for any $i \in [m]$,

$$\Pr_{(S,U) \sim D^{m+1}}(|\mathrm{R}_{S^i}(f_S) - \mathrm{R}_{S^i}(f_{S^{i,U}})| \leq \beta) \geq 1 - \delta .$$

Combining CV and overlap stability, Kutin and Niyogi arrive at the next definition.

*Definition 2.4:* A learning algorithm $\mathcal{A}$ is $(\beta,\delta)$-*training-stable* or, has *training stability* $(\beta,\delta)$, if $\mathcal{A}$ has

i. CV stability $(\beta,\delta)$; and
ii. Overlap stability $(\beta,\delta)$.

Trivially weak hypothesis stability $(\beta,\delta)$ implies training stability $(\beta,\delta)$. Kutin and Niyogi show that training stability is sufficient for good bounds on generalization error (for ERM, CV stability is sufficient). They also consider the stability of ERM on a two-classifier hypothesis class [5, Theorems 9.2 and 9.5], showing that the achievable rate on $\delta$ undergoes a phase transition as the number of risk minimizers increases from one to two.

*Theorem 2.5:* Consider the class $\mathcal{H}$ consisting of the two constant classifiers mapping $\mathcal{X}$ to $-1$ and $1$ respectively, and let $\mathcal{A}$ implement ERM on $\mathcal{H}$ (*i.e.*, $\mathcal{A}$ outputs a majority label of training set $S$). Let $p = \Pr_{(X,Y) \sim D}(Y=1)$.

1. If $p \geq \frac{1}{2}$ then $\mathcal{A}$ is $(0,\delta)$-training-stable for $\delta(m) \approx (2\pi m)^{-1/2}$.
2. If $p > \frac{1}{2}$ then $\mathcal{A}$ is weakly $(0,\delta)$-hypothesis-stable for
$$\delta = \exp\left(-\left(2-p^{-1}\right)^2 m/8 + \mathrm{O}(1)\right).$$
3. If $p = \frac{1}{2}$ then $\mathcal{A}$ is not $(\beta,\delta)$-CV-stable, or $(\beta,\delta)$-overlap-stable for any $\beta < 1$ and any $\delta = \mathrm{o}\left(m^{-1/2}\right)$.

The case of finite hypothesis spaces is then considered. In particular the authors show that fast rates are achieved in the presence of a unique risk minimizer.

*Theorem 2.6:* Let $\mathcal{H}$ be a finite collection of classifiers, and let $\mathcal{A}$ be a learning algorithm which performs ERM over $\mathcal{H}$. Suppose there exists a unique risk minimizer $h^\star \in \mathcal{H}$, then $\mathcal{A}$ is weakly $(0,\delta)$-hypothesis-stable for $\delta = \exp(-\Omega(m))$.

The authors then conjecture that the analogue of Theorem 2.5.(3) holds for general $|\mathcal{H}| < \infty$: [5, Conjecture 10.11] predicts that under any distribution $D$ inducing multiple risk minimizers, ERM is not $(0,\delta)$-training-stable for any $\delta = \mathrm{o}\left(m^{-1/2}\right)$.

## III. Finite Classes with Multiple Risk Minimizers

*Lemma 3.1:* Let $\mathcal{H}$ be a finite hypothesis space with $|\mathcal{H}| \geq 2$ and with risk minimizers satisfying $|\mathcal{H}^\star| = 2$. Then ERM on $\mathcal{H}$ with respect to the 0-1 loss, on a sample of $m$ examples, is not $(0,\delta)$-CV-stable for any $\delta = \mathrm{o}\left(m^{-1/2}\right)$. Furthermore, ERM on $\mathcal{H}$ is $(0,\delta)$-CV-stable for $\delta = \mathrm{O}\left(m^{-1/2}\right)$.

*Proof:* Let $\mathcal{H}^\star = \{h_1, h_2\}$. We proceed by first showing that $f_S$ lies in $\mathcal{H}^\star$ with exponentially increasing probability and then that switching within $\mathcal{H}^\star$ occurs often. Let $\epsilon = \min_{h \in \mathcal{H} \setminus \mathcal{H}^\star} \mathrm{R}_D(h) - \mathrm{R}_D(h_1)$, which exists and is positive by the finite cardinality of $\mathcal{H}$. Then by the union bound, $\mathrm{R}_D(h) \geq \mathrm{R}_D(h_1) + \epsilon$ for all $h \in \mathcal{H} \setminus \mathcal{H}^\star$, and Chernoff's bound

$$\begin{aligned}
&\Pr(f_S \in \mathcal{H}^\star) \\
&\geq \Pr(\forall h \in \mathcal{H} \setminus \mathcal{H}^\star, \mathrm{R}_S(h_1) < \mathrm{R}_S(h)) \\
&\geq \Pr(\forall h \in \mathcal{H} \setminus \mathcal{H}^\star, \mathrm{R}_S(h_1) < \mathrm{R}_D(h_1) + \epsilon/2 \leq \mathrm{R}_S(h)) \\
&\geq 1 - \Pr(\mathrm{R}_S(h_1) \geq \mathrm{R}_D(h_1) + \epsilon/2) \\
&\quad - \sum_{h \in \mathcal{H} \setminus \mathcal{H}^\star} \Pr(\mathrm{R}_D(h_1) + \epsilon/2 > \mathrm{R}_S(h)) \\
&\geq 1 - \Pr(\mathrm{R}_S(h_1) \geq \mathrm{R}_D(h_1) + \epsilon/2) \\
&\quad - \sum_{h \in \mathcal{H} \setminus \mathcal{H}^\star} \Pr(\mathrm{R}_D(h) - \epsilon/2 > \mathrm{R}_S(h)) \\
&\geq 1 - \exp\left(-\epsilon^2 m/2\right) - \sum_{h \in \mathcal{H} \setminus \mathcal{H}^\star} \exp\left(-\epsilon^2 m/2\right) \\
&= 1 - (|\mathcal{H}| - 1)\exp\left(-\epsilon^2 m/2\right) . \quad (1)
\end{aligned}$$

Consider ERM on $\mathcal{H}^\star$. Without loss of generality assume that ERM on $\mathcal{H}^\star$, when $h_1$ and $h_2$ have equal empirical risk, selects $h_1$. Let $\mathcal{Z}_1 = \{z \in \mathcal{Z} \mid \ell(h_1, z) < \ell(h_2, z)\}$ and $\mathcal{Z}_2$ be defined analogously. Let $p = \Pr(Z \in \mathcal{Z}_1 \cup \mathcal{Z}_2)$, which is positive by assumption that $h'(X) = h''(X)$ a.s. implies $h' = h''$. Then for all $i \in [m]$

$$\Pr_{D^2}((Z_i, U) \in (\mathcal{Z}_1 \times \mathcal{Z}_2) \cup (\mathcal{Z}_2 \times \mathcal{Z}_1)) = p^2/2 . \quad (2)$$

For the moment, assume that for all $i \in [m]$

$$\Pr\bigl(|\mathrm{R}_{S^i}(h_1) - \mathrm{R}_{S^i}(h_2)| \leq (m-1)^{-1}\bigr) = \mathrm{O}\left(m^{-1/2}\right) . \quad (3)$$

Conditioned on the events of (2) and (3), the probability of ERM on $S$ outputting a different hypothesis than on $S^{i,U}$ is at least $1/2$. This fact can be proved by cases on $|\mathrm{R}_{S^i}(h_1) - \mathrm{R}_{S^i}(h_2)|$, while assuming that $(Z_i, U) \in (\mathcal{Z}_1 \times \mathcal{Z}_2) \cup (\mathcal{Z}_2 \times \mathcal{Z}_1)$. If $\mathrm{R}_{S^i}(h_1) = \mathrm{R}_{S^i}(h_2)$, then $f_S \neq f_{S^{i,U}}$ trivially. If $|\mathrm{R}_{S^i}(h_1) - \mathrm{R}_{S^i}(h_2)| = (m-1)^{-1}$ then $\{|\mathrm{R}_{S^i,Z}(h_1) - \mathrm{R}_{S^i,Z}(h_2)| : Z \in \{Z_i, U\}\} = \{0, 2m^{-1}\}$ and it follows that $|\mathrm{R}_S(h_1) - \mathrm{R}_S(h_2)| \neq |\mathrm{R}_{S^{i,U}}(h_1) - \mathrm{R}_{S^{i,U}}(h_2)|$; and since by symmetry the probability that $\mathrm{R}_{S^i,Z}(h_2) < \mathrm{R}_{S^i,Z}(h_1)$ conditioned on the corresponding difference being $2m^{-1}$ is $1/2$, it follows that $f_S \neq f_{S^{i,U}}$ with probability $1/2$. Thus we have shown that on $\mathcal{H}^\star$, for all $i \in [m]$

$$\Pr_{(S,U) \sim D^{m+1}}(f_S \neq f_{S^{i,U}} \mid \mathscr{A} \cap \mathscr{B}) \geq 1/2 , \quad (4)$$

where

$$\begin{aligned}
\mathscr{A} &= \{(S,U) : (Z_i, U) \in (\mathcal{Z}_1 \times \mathcal{Z}_2) \cup (\mathcal{Z}_2 \times \mathcal{Z}_2)\} \\
\mathscr{B} &= \{(S,U) : |\mathrm{R}_{S^i}(h_1) - \mathrm{R}_{S^i}(h_2)| \leq (m-1)^{-1}\} .
\end{aligned}$$

Notice that since $f_S, f_{S^{i,U}} \subseteq \mathcal{H}^\star$ and $U \in \mathcal{Z}_1 \cup \mathcal{Z}_2$, $f_S \neq f_{S^{i,U}}$ implies that $\ell(f_S, U) \neq \ell(f_{S^{i,U}}, U)$. Then together (1)–(4) lead to the following statement about ERM over $\mathcal{H}$:

$$\begin{aligned}
&\Pr_{(S,U) \sim D^{m+1}}(|\ell(f_S, U) - \ell(f_{S^{i,U}}, U)| > 0) \\
&\geq \Pr(|\ell(f_S, U) - \ell(f_{S^{i,U}}, U)| > 0, \, f_S, f_{S^{i,U}} \in \mathcal{H}^\star) \\
&\geq \mathrm{O}\left(p^2 m^{-1/2}\right)\left(1 - 2(|\mathcal{H}|-1)\exp\left(-\epsilon^2 m/2\right)\right) \\
&= \mathrm{O}\left(m^{-1/2}\right) .
\end{aligned}$$

And so ERM is not $(0,\delta)$-CV-stable for any $\delta = o(m^{-1/2})$ as claimed. Furthermore ERM is $(0,\delta)$-CV-stable for $\delta = O(m^{-1/2})$ since

$$\begin{aligned}
&\Pr_{(S,U)\sim D^{m+1}}(|\ell(f_S, U) - \ell(f_{S^{i,U}}, U)| > 0) \\
&\leq \Pr(|\ell(f_S, U) - \ell(f_{S^{i,U}}, U)| > 0, \; f_S, f_{S^{i,U}} \in \mathcal{H}^\star) \\
&\quad + \Pr(f_S \notin \mathcal{H}^\star \vee f_{S^{i,U}} \notin \mathcal{H}^\star) \\
&\leq O(m^{-1/2}) + 2(|\mathcal{H}| - 1)\exp(-\epsilon^2 m/2) \\
&= O(m^{-1/2}) \; .
\end{aligned}$$

All that remains is to verify (3). Consider $X_{n,q} \sim \mathrm{Bin}(n,q)$ for $n \in \mathbb{N}, q \in [0,1]$ and note that for $k \in \mathbb{N} \cup \{0\}$

$$\begin{aligned}
\Pr\left(\left|X_{k,\frac{1}{2}} - \frac{k}{2}\right| \leq \frac{1}{2}\right) &= \begin{cases} \Pr\left(X_{k,\frac{1}{2}} = \frac{k}{2}\right), & k \text{ even} \\ 2\Pr\left(X_{k,\frac{1}{2}} = \frac{k-1}{2}\right), & k \text{ odd} \end{cases} \\
&\geq \Pr\left(X_{2\lfloor \frac{k}{2}\rfloor + 1, \frac{1}{2}} = \left\lfloor \frac{k}{2} \right\rfloor\right) \; .
\end{aligned}$$

Splitting on $\sum_{j=1: j\neq i}^m \mathbf{1}[Z_j \in \mathcal{Z}_1 \cup \mathcal{Z}_2]$—the $\mathrm{Bin}(m-1, p)$ number of examples in $S^i$ on which $h_1, h_2$ disagree—and noting that when conditioned on this sum being equal to $k$, the random variable $\frac{m-1}{2}(\mathrm{R}_{S^i}(h_1) - \mathrm{R}_{S^i}(h_2)) + \frac{k}{2} \sim \mathrm{Bin}(k, 1/2)$, leads to the lower-bound

$$\begin{aligned}
&\Pr(|\mathrm{R}_{S^i}(h_1) - \mathrm{R}_{S^i}(h_2)| \leq (m-1)^{-1}) \\
&= \sum_{k=0}^{m-1} \Pr(X_{m-1,p} = k) \Pr\left(\left|X_{k,\frac{1}{2}} - \frac{k}{2}\right| \leq \frac{1}{2}\right) \\
&\geq \min_{0\leq k\leq m-1} \Pr\left(\left|X_{k,\frac{1}{2}} - \frac{k}{2}\right| \leq \frac{1}{2}\right) \sum_{j=0}^{m-1} \Pr(X_{m-1,p} = j) \\
&= \min_{0\leq k\leq m-1} \Pr\left(\left|X_{k,\frac{1}{2}} - \frac{k}{2}\right| \leq \frac{1}{2}\right) \\
&\geq \Pr\left(X_{2\lfloor \frac{m}{2}\rfloor + 1, \frac{1}{2}} = \left\lfloor \frac{m}{2} \right\rfloor\right) \\
&= O(m^{-1/2}) \; ,
\end{aligned}$$

where the last relation is a consequence of Stirling's approximation. Let $c = \lceil p(m-1)/2 \rceil$. The upper-bound follows similarly:

$$\begin{aligned}
&\Pr(|\mathrm{R}_{S^i}(h_1) - \mathrm{R}_{S^i}(h_2)| \leq (m-1)^{-1}) \\
&= \sum_{k=0}^{c} \Pr(X_{m-1,p} = k) \Pr\left(\left|X_{k,\frac{1}{2}} - \frac{k}{2}\right| \leq \frac{1}{2}\right) \\
&\quad + \sum_{k=c+1}^{m-1} \Pr(X_{m-1,p} = k) \Pr\left(\left|X_{k,\frac{1}{2}} - \frac{k}{2}\right| \leq \frac{1}{2}\right) \\
&\leq \Pr(X_{m-1,p} \leq c) \\
&\quad + \sup_{k \in \{c+1,\ldots,m-1\}} \Pr\left(\left|X_{k,\frac{1}{2}} - \frac{k}{2}\right| \leq \frac{1}{2}\right) \\
&\leq O(e^{-pm}) + O(m^{-1/2}) \\
&= O(m^{-1/2}) \; ,
\end{aligned}$$

where the penultimate relation follows from an application of Chernoff's inequality to the first term and Stirling's approximation to the second. $\square$

*Theorem 3.2:* Let $\mathcal{H}$ be a finite hypothesis space with $|\mathcal{H}| \geq 2$ and with risk minimizers $|\mathcal{H}^\star| > 1$. Then any learning algorithm $\mathcal{A}$ implementing ERM on $\mathcal{H}$ with respect to the 0-1 loss, on a sample of $m$ examples, is not $(0,\delta)$-CV-stable for any $\delta = o(m^{-1/2})$. Furthermore, $\mathcal{A}$ is $(0,\delta)$-CV-stable for $\delta = O(m^{-1/2})$.

*Proof:* Define $\epsilon$ as before. Arbitrarily order the risk minimizers $\mathcal{H}^\star = \{h_1, \ldots, h_n\}$ where $n = |\mathcal{H}^\star|$, and let $p_{ij} = \Pr(h_i(X) \neq h_j(X)) > 0$ by assumption that $h'(X) = h''(X)$ a.s. implies $h' = h''$ for each $h', h'' \in \mathcal{H}$. Let $f_T^{ij}$ be the result of running $\mathcal{A}$ on sample $T$ in the hypothesis space $\{h_i, h_j\}$, where ties between empirical risk minimizers are broken as in ERM over $\mathcal{H}^\star$. Let $f_T^\star$ be the result of running $\mathcal{A}$ on $T$ in $\mathcal{H}^\star$. It follows that for all $\{j,k\} \subseteq [n]$, conditional upon event $\mathscr{C}_{jk} = \{f_S^\star = h_j, f_{S^{i,U}}^\star = h_k\}$, $f_S^{jk} = f_S^\star = h_j$ and $f_{S^{i,U}}^{jk} = f_{S^{i,U}}^\star = h_k$, and so by (2)–(4)

$$\begin{aligned}
&\Pr_{(S,U)\sim D^{m+1}}(|\ell(f_S^\star, U) - \ell(f_{S^{i,U}}^\star, U)| > 0 \mid \mathscr{C}_{jk}) \\
&= \Pr\left(\left|\ell\left(f_S^{jk}, U\right) - \ell\left(f_{S^{i,U}}^{jk}, U\right)\right| > 0 \;\middle|\; \mathscr{C}_{jk}\right) \\
&\geq O\left(\min_{\{j,k\}\subseteq [n]} p_{jk}^2 m^{-1/2}\right) \; . \quad (5)
\end{aligned}$$

It follows that

$$\begin{aligned}
&\Pr_{(S,U)\sim D^{m+1}}(|\ell(f_S^\star, U) - \ell(f_{S^{i,U}}^\star, U)| > 0) \\
&\geq \min_{\{j,k\}\subseteq [n]} \Pr(|\ell(f_S^\star, U) - \ell(f_{S^{i,U}}^\star, U)| > 0 \mid \mathscr{C}_{jk}) \\
&\geq O\left(\min_{\{j,k\}\subseteq [n]} p_{jk}^2 m^{-1/2}\right) \; . \quad (6)
\end{aligned}$$

Observing that $\Pr(f_S \in \mathcal{H}^\star) \geq 1 - |\mathcal{H}|\exp(-\epsilon^2 m/2)$ by the same argument that lead to (1), we have as before

$$\begin{aligned}
&\Pr(|\ell(f_S, U) - \ell(f_{S^{i,U}}, U)| > 0) \\
&\geq \Pr(|\ell(f_S, U) - \ell(f_{S^{i,U}}, U)| > 0, \; f_S, f_{S^{i,U}} \in \mathcal{H}^\star) \\
&\geq \Pr(|\ell(f_S^\star, U) - \ell(f_{S^{i,U}}^\star, U)| > 0) \\
&\quad \times (1 - \Pr(f_S \notin \mathcal{H}^\star \vee f_{S^{i,U}} \notin \mathcal{H}^\star)) \\
&= O\left(\min_{\{j,k\}\subseteq [n]} p_{jk}^2 m^{-1/2}\right)(1 - 2|\mathcal{H}|\exp(-\epsilon^2 m/2)) \\
&= O(m^{-1/2}) \; . \quad (7)
\end{aligned}$$

This implies that ERM on $\mathcal{H}$ is not $(0,\delta)$-CV-stable for any $\delta = o(m^{-1/2})$. Similarly we derive an upper-bound analogous to (7) by the same technique used in the proof of Lemma 3.1:

$$\begin{aligned}
&\Pr(|\ell(f_S, U) - \ell(f_{S^{i,U}}, U)| > 0) \\
&\leq \Pr(|\ell(f_{S^{i,U}}, U) - \ell(f_S, U)| > 0, \; f_S, f_{S^{i,U}} \in \mathcal{H}^\star) \\
&\quad + \Pr(f_S \notin \mathcal{H}^\star \vee f_{S^{i,U}} \notin \mathcal{H}^\star) \\
&= O\left(\max_{\{j,k\}\subseteq [n]} p_{jk}^2 m^{-1/2}\right) + 2|\mathcal{H}|\exp(-\epsilon^2 m/2) \\
&= O(m^{-1/2}) \; .
\end{aligned}$$

This implies that ERM on $\mathcal{H}$ is $(0,\delta)$-CV-stable for $\delta = O(m^{-1/2})$. $\square$

ACKNOWLEDGMENTS

The authors thank Peter Bartlett for his helpful feedback on this research.

**Benjamin I. P. Rubinstein** received the BSc and BE degrees in Pure Mathematics and Software Engineering respectively from the University of Melbourne, Australia, in 2003 and the MCompSci degree from the same university in 2009.

He has been a PhD student in the Department of Electrical Engineering and Computer Sciences, University of California at Berkeley since 2004, advised by Prof. Peter Bartlett. He has worked at Google Research, Yahoo! Research and Intel Research Berkeley. His research interests are in applications of machine learning to measurement, computer security and privacy, and in the fundamental limits of learning in adversarial environments.

**Aleksandr Simma** received the BS degree in Computer Engineering and a BS in Management Science from the University of California at San Diego, in 2004.

Since then, he has been a PhD student in the Department of Electrical Engineering and Computer Sciences, University of California at Berkeley, advised by Prof. Michael Jordan. His research interests include point processes and applications of learning approaches in finance.